\documentclass[11pt,a4paper]{article}

\newenvironment{arxivabstract}

\usepackage[a4paper,margin=1in]{geometry}
\usepackage[T1]{fontenc}
\usepackage[utf8]{inputenc}
\usepackage{lmodern}
\usepackage[authoryear]{natbib}
\usepackage{authblk}
\usepackage{graphicx}
\usepackage{booktabs}
\usepackage{float}
\usepackage{multirow}
\usepackage{array}
\usepackage{algorithm}
\usepackage{algpseudocode}
\usepackage{subcaption}
\usepackage{amsmath,amssymb}
\usepackage{enumitem}
\usepackage{float}
\usepackage{booktabs}
\usepackage{graphicx}
\usepackage[hidelinks]{hyperref}

\begin{document}

\title{Sparse Visual Aggregation and
Direction-Aware Temporal Encoding for
Zero-Shot Neural Visual Decoding}

\author[1]{Neha}
\author[1]{Mudasir Ganaie}

\affil[1]{Department of Computer Science and Engineering,
Indian Institute of Technology Ropar,
Rupnagar, Punjab 140001, India}

\date{}
\maketitle
\begin{arxivabstract}
Brain–computer interfaces aim to establish direct communication between neural activity and external systems, making reliable interpretation of brain signals an important engineering problem. In this context, neural visual decoding seeks to infer visual information from neural activity by learning a mapping between brain responses and the corresponding visual stimuli. Our brain primarily follows a hierarchical pathway to process visual information captured by the eye. However, neural representations learned by existing methods differ fundamentally from those produced by the biological visual system. To bridge this gap and mimic the human vision hierarchy, we introduce Sparse Visual Aggregation and Direction-Aware Temporal Encoding for Zero-Shot Neural Visual Decoding. This framework aligns neural visual representations by jointly modeling early visual processes via fovea- and multi-scale difference-of-Gaussians transformations. To align the semantic information with neural information, it extracts multiple intermediate representations of the image at multiple depths and passes them through the Residual Gated Adapter Mechanism. To study the temporal dynamics of neural signals, the Direction-aware Multi-Scale Temporal Attention Encoder is introduced to align the learned neural representation with the final visual representation. The proposed framework achieves 92.5\% top-1 and 99.27\% top-5 accuracy in the intra-subject setting on the THINGS-EEG dataset, and 50.25\%  and 75\% top-1 and top-5 accuracy in the THINGS-MEG dataset, respectively. These results indicate that combining biologically motivated visual representations with the structured modalities of neural dynamics improves cross-modal alignment and zero-shot image retrieval.
\end{arxivabstract}

\section{Introduction}
The ability to interpret brain activity has become a primary aim in brain-computer interface development, enabling direct interaction between humans and external systems without relying on muscular activity~\citep{CHEN2025110654, LIAN2026116089}. In this field, a key research direction is visual neural decoding, which aims to infer the visual stimulus perceived by a subject from recorded neural signals. Most studies rely on electroencephalography (EEG) or magnetoencephalography (MEG), as both techniques record neural activity at millisecond resolution and are well suited for capturing the fast dynamics of visual processing. Nevertheless, EEG and MEG signals have low signal-to-noise ratio and exhibit substantial variability across and within subjects, making robust visual representation learning from neural activity a challenging problem ~\citep{BIASIUCCI2019R80,singh2014magnetoencephalography,ferrante2025towards}. These challenges are particularly pronounced in zero-shot visual retrieval, where the model must identify the image corresponding to a given neural signal from a mutually exclusive candidate set of images. Despite differences in sensing modalities, existing methods shares a common objective of recovering visual information from brain activity through tasks such as image retrieval ~\citep{song2024decoding,du2023decoding,zhang2026neurobridge}, image generation ~\citep{kavasidis2017brain2image,rathore2026multi}, caption generation ~\cite{mishra2025thought2text}, object recognition ~\cite{alazrai2022deep} and visual reconstruction ~\citep{bai2023dreamdiffusion,lopez2025guess,lan2023seeing, QU2026131526}. 

Recent studies have explored the visual side of neural–visual alignment through biologically motivated image transformations, feature degradation, evidence-based weighting, and intermediate-layer representations ~\citep{wu2025bridging,wu2026shrinking,shao2026braininspiredcaptureevidencedrivenneuromimetic}, demonstrating that visual representations can substantially influence decoding performance. However, existing approaches primarily investigate these strategies independently and ignore biologically inspired visual processing, adaptive refinement of pretrained representations, and hierarchical fusion across multiple representation levels. First, although biologically motivated transformations have been explored, the complementary information captured by distinct visual processing pathways remains unexplored within a pretrained vision encoder. In particular, early visual processing in the human visual system involves center--surround filtering and the segregation of information into parallel pathways before signals reach the visual cortex ~\citep{Hubel1962ReceptiveFB, Dacey2000ParallelPF}. Second, pretrained vision encoders are commonly used without explicitly adapting their intermediate representations to the neural signals, leaving cross-modal discrepancy between neural and visual representations. Finally, existing approaches rely on final layers or predefined feature aggregation strategies, which may not fully exploit the hierarchical representations available within pretrained vision models.
 
To overcome these limitations, a unified framework for zero-shot visual retrieval from EEG and MEG signals. It passes the transformed image representations to a pretrained vision model. It extracts multiple intermediate inter-layer feature representations, followed by a residual gated adapter mechanism and a fusion block for semantic representation. A Direction-aware multi-scale neural encoder provides the neural representation, which aligns with the learnable representation of visual stimuli in a shared embedding space through a contrastive learning paradigm.

The main contributions of the proposed framework are summarized as follows:
\begin{itemize}

 \item A multi-scale visual processing strategy that explores fovea filtering with the multi-scale Difference-of-Gaussians (DoG) processing to emphasize human hierarchical vision perception.

\item A sparse inter-layer residual aggregation mechanism that integrates representations from selected intermediate stages of a pretrained vision backbone to support neural–visual alignment.

\item A Residual Gated Refinement Adapter Module (ReGRAM) and Direction-aware Multi-Scale Temporal Attention Encoder (DMSTAE) encoder for the vision and brain representations, respectively, enabling more discriminative neural representations across multiple temporal scales.
\end{itemize}

\section{Related Work}

Recent advances in neural--visual decoding have explored several directions to bridge the representational gap between brain signals and visual information. These efforts can be broadly viewed from four perspectives: (i) modeling the temporal dynamics of EEG/MEG signals to obtain discriminative neural representations,  (ii) constructing hierarchical and biologically inspired visual representations that better reflect the structure of visual processing,(iii) employing pretrained visual encoders and lightweight adaptation mechanisms to improve the compatibility of visual features with neural representations, and (iv) learning cross-modal correspondence through alignment objectives such as contrastive learning. The following discussion reviews these directions and highlights the limitations that motivate our approach.

Temporal representation learning is central to extracting discriminative neural features from EEG/MEG signals, as neural responses evolve following visual stimulation. A common design is to extract temporal features using temporal convolutions, followed by temporal downsampling before passing the resulting representations to a transformer ~\citep{spampinato2017deep, palazzo2020decoding}. These approaches primarily capture temporal patterns through local convolutions and subsequent temporal aggregation. NICE ~\cite{song2024decoding} introduces a Temporal-Spatial Convolution (TSConv) module to learn richer neural features, while STAMBRIDGE ~\cite{meng2026stambridge} combines multi-scale temporal convolutions with spectral-temporal amplitude-aware modulation. Adaptive Teaching System (ATS) ~\cite{wu2026shrinking} introduces a Shared Temporal Attention Encoder to reduce the representation gap between EEG and visual features, whereas NeuroAlign ~\cite{ke2026neuroalign} separates perceptual and cognitive representations. However, the evolution of EEG signals depends not only on the magnitude of the response but also on how the signal changes over time. Standard temporal convolutions can capture local patterns, yet they ignore differences between opposite directions of temporal change. Aforementioned studies improve temporal representation learning from different perspectives, they do not explicitly model the direction of temporal neural transitions as an independent source of information.

In order to mimic human eye behavior, biologically inspired visual representations have been explored to improve neural–visual alignment.  Uncertainty-aware Blur Prior ~\cite{wu2025bridging} performs a fovea blurring technique dynamically to match the coarseness of neural representations, while NeuroBridge ~\cite{zhang2026neurobridge} applies Gaussian blur, noise, and mosaic effects to simulate perceptual variability. BI-Cap ~\cite{shao2026braininspiredcaptureevidencedrivenneuromimetic} goes further with neuromimetic transformations that approximate biological visual processing. 

Many visual neural decoding frameworks employ pretrained vision encoders such as CLIP ~\cite{radford2021learning}, ResNet ~\cite{he2016deep}, DINOv2 ~\cite{oquab2023dinov2}, and ViT ~\cite{dosovitskiy2021an}, which often keep the backbone frozen. Representation methods such as NICE ~\cite{song2024decoding} and BraVL ~\cite{du2023decoding} rely on pretrained visual representations and place most of the learning focus within the EEG encoder and the cross-modal projection network. HMAVD ~\cite{sun2025decoding}, ATS ~\cite{wu2026shrinking}, STAMBRIDGE ~\cite{meng2026stambridge}, BI-Cap ~\cite{shao2026braininspiredcaptureevidencedrivenneuromimetic} introduce an adapter mechanism, intermediate semantic spaces, or adaptive feature weighting to improve neural-visual alignment. Shallow alignment ~\cite{du2026deep} showed that aligning EEG representations with an intermediate visual layer outperforms using the final output, as deeper layers increasingly emphasize semantic invariance at the expense of low-level visual details. HCF ~\cite{tang2026aligning} jointly combines visual representations from different layers to better align with the multi-stage nature of human visual processing. RAHS ~\cite{zhu2026region} performs region-aware hierarchical sub-feature alignment, and SAMGA ~\cite{jiang2026subjectawaremultigranularityalignmentzeroshot} addresses the subject-dependent aspect of granularity selection.

Beyond the choice of visual representations, the objective used to establish correspondence between neural and visual modalities is also crucial. Contrastive learning has therefore become a common strategy for aligning heterogeneous modalities by bringing matched representations closer while separating mismatched ones in a shared embedding space ~\citep{QUAN2024102447, MAI2023101920, ZHONG2026104156}. This formulation has also been adopted in neural--visual decoding, where EEG or other brain representations are aligned with visual features for cross-modal retrieval using bidirectional contrastive loss ~\citep{wu2026shrinking, du2026deep}. Despite these advances, existing approaches primarily focus on establishing correspondence between the resulting neural and visual embeddings. At the same time, comparatively less attention is given to constructing and refining the hierarchical visual representation prior to cross-modal alignment. Motivated by this gap, our framework learns richer hierarchical visual representations through adaptive inter-layer aggregation and refinement, and subsequently aligns them with M/EEG representations using a contrastive objective.

\section{Proposed Sparse Inter-layer Residual Aggregation framework}
\begin{figure*}
  \centering
 \includegraphics[width=1.0\textwidth, height=1.0\textheight,keepaspectratio]{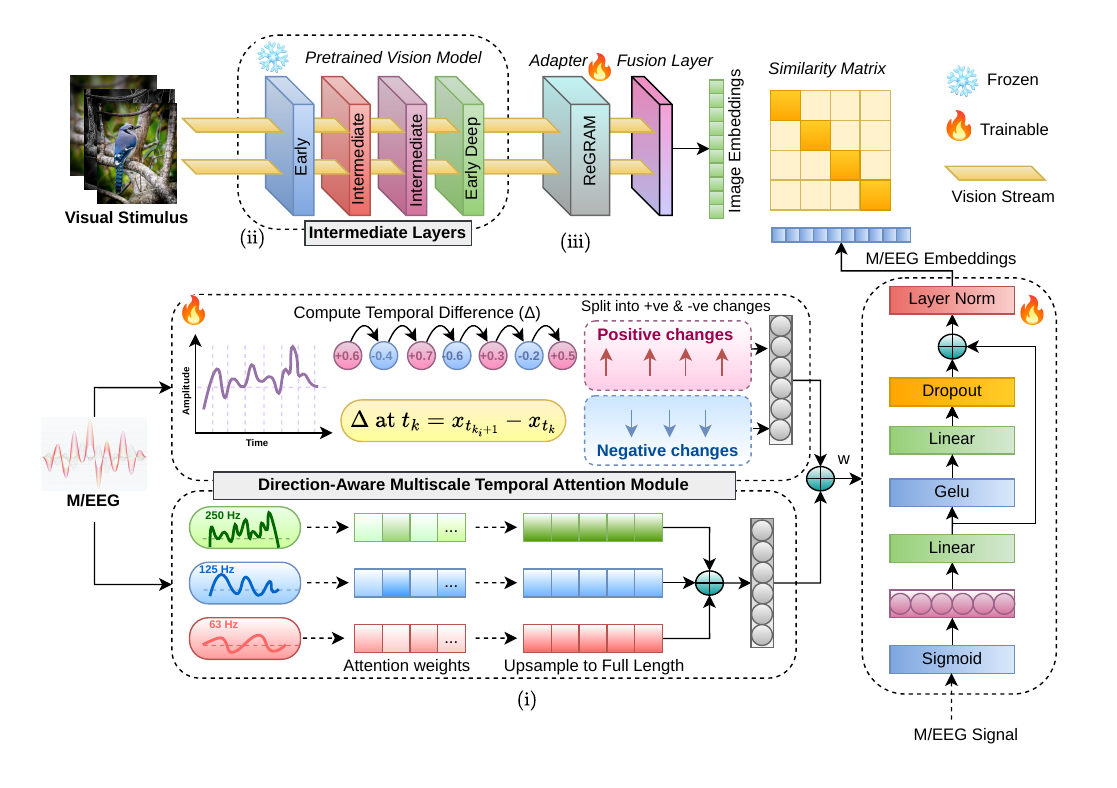}
  \caption{Overview of the proposed architecture. (i) Direction-aware Multi-Scale Temporal Attention Encoder for M/EEG representation learning, (ii) hierarchical intermediate-layer representation learning from a pretrained vision model, and (iii)Residual Gated Refinement Adapter Module (ReGRAM).}
  \label{fig:Proposed_architecture_diagram}
\end{figure*}

An overview of the proposed architecture is presented in Figure~\ref{fig:Proposed_architecture_diagram}. The framework consists of a Direction-Aware Multiscale Temporal Attention Encoder for M/EEG representation learning, hierarchical intermediate-layer visual representation learning, and a Residual Gated Refinement Adapter (ReGRAM) for cross-modal alignment.

\subsection{Problem Definition}
 Given a training dataset of $N$ paired EEG-image samples,
$\mathcal{D}=\{(\mathbf{X}_i,\mathbf{I}_i)\}_{i=1}^{N},$ 
where $i$ identifies the \(i\)-th EEG–image pair in the dataset, $\mathbf{X}_i \in \mathbb{R}^{C \times T}$ denotes the EEG signal with $C$ recording channels and $T$ temporal samples. And $\mathbf{I}_i\in\mathbb{R}^{H\times W\times 3}$ denotes the corresponding RGB image, where $H$ and $W$ represent the image height and width, respectively. Each image is first transformed into two complementary visual streams,
\begin{equation*}
\mathbf{I}^{F}_i=\mathcal{T}_{F}(\mathbf{I}_i),
\qquad
\mathbf{I}^{M}_i=\mathcal{T}_{M}(\mathbf{I}_i), \nonumber
\end{equation*}
where $\mathcal{T}_{F}$ and $\mathcal{T}_{M}$ denote the foveated and
multi-scale Difference-of-Gaussians transformations, respectively. A pretrained visual backbone is used to extract hierarchical visual representations. The resulting intermediate layer representations are adapted and aggregated to obtain the visual embedding given by 
\begin{equation*}
\mathbf{z}_{v,i}=f_v(\mathbf{I}_i^{F},\mathbf{I}_i^{M}), \nonumber
\end{equation*}
where $f_v(\cdot)$ denotes the visual encoding function that includes the
shared vision backbone, hierarchical feature aggregation, and visual
feature adaptation. Similarly, the neural signal is encoded as
\begin{equation*}
\mathbf{z}_{e,i}=f_e(\mathbf{X}_i), \nonumber
\end{equation*}
where $f_e(\cdot)$ denotes the neural encoding function that maps the EEG
signal to the shared latent representation. The resulting neural and visual embedding representations over a batch of paired samples are denoted by $\mathbf{Z}_{e}$ and $\mathbf{Z}_{v}$.
The two representations are jointly optimized in the shared latent space
using a bidirectional contrastive objective:
\begin{equation*}
\min_{\Theta}\;
\mathcal{L}_{\mathrm{CLIP}}
\left(\mathbf{Z}_{e},\mathbf{Z}_{v}\right), \qquad \mathbf{Z}_{e},\mathbf{Z}_{v}\in\mathbb{R}^{B\times D} \nonumber
\nonumber
\end{equation*}
where $\Theta$ denotes the trainable parameters of the framework, with $B$ denoting the batch size and $D$ the shared embedding dimension.

\subsection{Direction-aware Multi-Scale Temporal Attention Encoder}
The temporal dynamics of neural signals provide important information for distinguishing neural responses to visual stimuli. However, these dynamics can involve temporal changes occurring at different time scales and in different directions. To capture informative temporal patterns at multiple resolutions while distinguishing the direction of temporal changes, we introduce the Direction-aware Multi-Scale Temporal Attention Encoder (DMSTAE). Given a neural signal $\mathbf{X}_i\in\mathbb{R}^{C\times T}$ for the $i$-th sample, the temporal difference at channel $c$ and time point $t$ is defined as
\begin{equation*}
\Delta X_{i,c,t}
=
X_{i,c,t+1}-X_{i,c,t}    \qquad\text{where }  \ t=1,\ldots,T-1 \nonumber
\qquad
\end{equation*}
The temporal differences are decomposed into positive and negative components:
\begin{equation*}
\Delta X^{+}_{i,c,t}
=
\max(\Delta X_{i,c,t},0),
\qquad
\Delta X^{-}_{i,c,t}
=
\max(-\Delta X_{i,c,t},0) \nonumber
\end{equation*}
This decomposition preserves the direction of temporal transitions, allowing increasing and decreasing neural responses to contribute independently to the subsequent attention mechanism. To account for the relative contribution of different neural channels, learnable channel weights are normalized separately for positive and negative temporal changes:
\begin{equation*}
\mathbf{p}
=
\operatorname{softmax}(\boldsymbol{\theta}^{+}),
\qquad
\mathbf{n}
=
\operatorname{softmax}(\boldsymbol{\theta}^{-}),
\qquad
\mathbf{p},\mathbf{n}\in\mathbb{R}^{C}, \nonumber
\end{equation*}
where $\boldsymbol{\theta}^{+},\boldsymbol{\theta}^{-}\in\mathbb{R}^{C}$ are learnable channel-weight parameters. The corresponding directional temporal energies are computed over a mini-batch of $B$ samples as
\begin{equation*}
e_t^{+}=\mathbb{E}_{i\sim\mathcal{B}}\left[\mathbf{p}^{\top}\Delta\mathbf{X}^{+}_{i,t}\right],
\qquad
e_t^{-}=\mathbb{E}_{i\sim\mathcal{B}}\left[\mathbf{n}^{\top}\Delta\mathbf{X}^{-}_{i,t}\right] \nonumber
\end{equation*}
where $\Delta\mathbf{X}^{+}_{i,t}$, $\Delta\mathbf{X}^{-}_{i,t}\in\mathbb{R}^{C}$ denote the vectors of positive and negative temporal changes, respectively, across the $C$ neural channels at time $t$. Here, $e_t^{+}$ and $e_t^{-}$ denote the batch-averaged positive and negative temporal energies at time $t$.

To capture temporal dependencies at multiple resolutions, three learnable
temporal attention patterns are defined at the original, half, and quarter
temporal resolutions:
\begin{equation*}
\mathbf{a}^{(1)}\in\mathbb{R}^{T},
\qquad
\mathbf{a}^{(2)}\in\mathbb{R}^{T/2},
\qquad
\mathbf{a}^{(3)}\in
\mathbb{R}^{\lceil T/4\rceil} \nonumber
\end{equation*}
The lower-resolution patterns are linearly interpolated to $T$ temporal
samples, yielding $\widetilde{\mathbf a}^{(2)}$, $\widetilde{\mathbf a}^{(3)}\in\mathbb{R}^{T}$. The three patterns provide complementary temporal views, with the original resolution preserving fine-grained temporal variations and the lower resolutions capturing progressively broader temporal context. Joint temporal resolution is defined as
\begin{equation*}
\mathbf{a}
=
\alpha_1\mathbf{a}^{(1)}
+
\alpha_2\widetilde{\mathbf{a}}^{(2)}
+
\alpha_3\widetilde{\mathbf{a}}^{(3)}, \nonumber
\end{equation*}
where $\alpha_1,\alpha_2,\alpha_3$ are learnable scalar coefficients. The directional energies modulate the multi-scale temporal pattern to produce the final temporal attention weights:
\begin{equation*}
\boldsymbol{\omega}
=
\operatorname{softmax}
\left(
\mathbf{a}
+
\gamma_{+}\mathbf{e}^{+}
+
\gamma_{-}\mathbf{e}^{-}
\right),
\qquad
\boldsymbol{\omega}\in\mathbb{R}^{T}, \nonumber
\end{equation*}
where $\mathbf e^{+},\mathbf e^{-}\in\mathbb{R}^{T}$ contain the
batch-averaged positive and negative temporal energies, respectively, and
$\gamma_{+}$ and $\gamma_{-}$ are learnable scalar coefficients controlling
respective contributions. The resulting weights provide a temporal importance profile that combines the multi-scale temporal patterns with the direction-aware neural responses. The attention weights are applied uniformly across the channels of each sample:
\begin{equation*}
\widetilde{\textbf{X}}_{i}
=
\omega_t X_{i,c,t},
\qquad
\widetilde{\mathbf X}_i\in\mathbb{R}^{C\times T} \nonumber
\end{equation*}
This operation emphasizes temporally informative responses while preserving the channel structure of the original neural signal. The resulting attended signal is flattened and projected into the shared $D$-dimensional embedding space:
\begin{equation*}
\mathbf{h}_i
=
\mathbf{W}_{p}
\operatorname{flatten}(\widetilde{\mathbf X}_i)
+
\mathbf{b}_{p},
\qquad
\mathbf{h}_i\in\mathbb{R}^{D}, \nonumber
\end{equation*}
where
$\mathbf{W}_{p}\in\mathbb{R}^{D\times CT}$ and
$\mathbf{b}_{p}\in\mathbb{R}^{D}$ are the projection weight matrix and bias.

Finally, a residual projection block follows from Wu et. al. ~\cite{wu2026shrinking}, refining the representation:
\begin{equation*}
\mathbf{z}_{e,i}
=
\operatorname{LN}
\left[
\mathbf{h}_i
+
\operatorname{Dropout}
\left(
\mathbf{W}_{2}
\operatorname{GELU}(\mathbf{h}_i)
+
\mathbf{b}_{2}
\right)
\right],
\qquad
\mathbf{z}_{e,i}\in\mathbb{R}^{D}, \nonumber
\end{equation*}
where $\mathbf{W}_{2}\in\mathbb{R}^{D\times D}$ and
$\mathbf{b}_{2}\in\mathbb{R}^{D}$ are the learnable parameters. The residual connection preserves the original projected representation while allowing the nonlinear transformation to refine its features. Layer normalization is then applied to obtain the final neural representation $\mathbf{z}_{e,i}$ for subsequent cross-modal alignment.

\subsection{Biologically Inspired Visual Representation Learning}

Each input image is transformed into two complementary visual streams: a foveated stream that preserves high-resolution information and a multi-scale DoG transformation that emphasizes luminance transitions and coarse spatial structure. Together, these complementary representations provide visual features that are consistent with the characteristics of human visual perception and neural responses.

\subsubsection{Foveated Visual Stream}
To model the non-uniform spatial resolution of human vision, we construct a foveated visual representation in which spatial details are preserved near the fixation region and progressively reduced toward the image periphery. 
Following the fovea imaging approach, the spatial importance at pixel location $(i,j)$ in an image is defined as
\begin{equation*}
A_F(i,j)
=
R_e+(R_c-R_e)
\exp\left(
-g\frac{d(i,j)}{d_{\max}}
\right), \nonumber
\end{equation*}
where $R_c$ and $R_e$ denote the central and peripheral spatial resolutions, respectively, $g$ controls the rate of spatial degradation, and $d(i,j)$ denotes the distance of pixel $(i,j)$ from the fixation center. $d_{\max}$ is the maximum distance from the fixation center within the
image. A Gaussian-blurred version of the input image is generated, and the foveated visual representation is constructed by combining the original and blurred images according to the spatial importance mask:
\begin{equation*}
\mathbf{I}^{F}
=
\mathbf{A}_F\odot\mathbf{I}
+
(\mathbf{1}-\mathbf{A}_F)\odot
\mathbf{I}_{\mathrm{blur}},
\qquad
\mathbf{I}^{F}\in\mathbb{R}^{H\times W\times3}, \nonumber
\end{equation*} 
where $\odot$ denotes element-wise multiplication and $\mathbf{1}$ is an all-ones mask with the same spatial dimensions as $\mathbf{A}_{F}$. The mask assigns larger values near the fixation center and smaller values toward the periphery. This results in greater preservation of visual details around the fixation region and progressively stronger smoothing toward the peripheral regions.

\subsubsection{multi-scale Difference-of-Gaussians Visual Stream}
To approximate the luminance-sensitive and coarse spatial processing associated with the magnocellular pathway of the human visual system, a luminance-based representation is constructed to emphasize local spatial contrast and suppress color-specific information.
Given an image $\mathbf{I}\in\mathbb{R}^{H\times W\times3}$, the luminance of the image is computed as
\begin{equation*}
L = w_R R + w_G G + w_B B,  \nonumber
\end{equation*}
where $R$, $G$, and $B$ denote the red, green, and blue channels, respectively, and $w_R$, $w_G$, and $w_B$ are the corresponding channel weights. We compute the DoG response at several spatial scales by subtracting a more strongly smoothed version of the luminance image from a less smoothed one given by:
\begin{equation*}
\mathbf{D}^{(k)}
=
G_{\sigma_{k,1}}(\mathbf{L})
-
G_{\sigma_{k,2}}(\mathbf{L}) \qquad \text{ and } \qquad \sigma_{k,2}>\sigma_{k,1},
\qquad
k\in\{1,2, \cdots n\},  \nonumber
\end{equation*}
where $G_{\sigma}(\cdot)$ denotes Gaussian smoothing with standard deviation $\sigma$, and $\mathbf{D}^{(k)}\in\mathbb{R}^{H\times W}$ denotes the DoG response at the $k$-th spatial scale. The response from the n scales is combined via a weighted combination:
\begin{equation*}
\mathbf{I}^{M}
=
\sum_{k=1}^{n}
w_k\mathbf{D}^{(k)}, \nonumber
\end{equation*}
where $w_k$ is a learnable scalar weight controlling the contribution of the $k$-th spatial scale.

\subsection{Intermediate Layer Representation}

Visual information is represented hierarchically in the brain, with neural responses measured by EEG and MEG reflecting different stages of visual processing. The hierarchical nature motivates the use of visual representations at multiple levels rather than relying solely on the final semantic representation of the vision transformer. Earlier and intermediate layers retain lower- and mid-level visual structures, while deeper layers capture increasingly abstract semantic information. Aligning neural responses with only the final representation may therefore discard visual information that is relevant to the EEG and MEG signals. To preserve these complementary levels of visual information, we extract features from progressively deeper stages of the pretrained vision transformer:
\begin{equation*}
\mathbf{H}^{s,k}
=
\left\{
\mathbf{I}^{s,k}
\right\}_{k\in\mathcal{K}},
\qquad s\in\{F,M\}, \nonumber
\end{equation*}
where $\mathcal{K}$ denotes the set of selected layers, $F$ and $M$ are fovea- and multi-scale DoG streams, respectively. For each visual stream, the features are progressively accumulated through residual connections:
\begin{equation*}
\mathbf{H}^{s,k_1}=\mathbf{I}^{s,k_1},\qquad
\mathbf{H}^{s,k}
=
\mathbf{I}^{s,k}
+
\mathbf{H}^{s,k^-},
\quad k\in\mathcal{K}\setminus\{k_1\}, \nonumber
\end{equation*}
where $k_1$ denotes the first selected layer and $k^-$ denotes the preceding selected layer. The residual accumulation preserves information from earlier stages while incorporating increasingly abstract representations from deeper layers. The resulting hierarchical representations are combined via learnable layer weights:
\begin{equation*}
\mathbf{T}^{s}
=
\sum_{k\in\mathcal{K}}
\beta_k^{s}\mathbf{H}^{s,k},
\qquad
\boldsymbol{\beta}^{s}
=
\operatorname{softmax}(\boldsymbol{\lambda}^{s}), \nonumber
\end{equation*}
where $\boldsymbol{\lambda}^{s}$ denotes the learnable layer-selection parameters. This allows the model to adaptively emphasize the visual depths that are informative for aligning the visual representation with EEG and MEG responses, producing the hierarchical representations $T_F$ and $T_M$ for the foveated and multi-scale DoG transformation streams, respectively.

\subsection{Residual Gated Refinement Adapter Module}

The hierarchical visual representations capture complementary information at different levels of abstraction; however, these representations require further refinement to better accommodate EEG and MEG alignment. Hence, introduce a lightweight residual gated adapter to selectively refine the visual features while preserving the information of the pretrained vision backbone.

For a stream representation $\mathbf{T}^{s}$, $s\in\{F,M\}$, the feature gate is computed as
\begin{equation*}
\mathbf{g}^{s}
=
\tanh
\left(
\mathbf{W}^{s}_{2}
\operatorname{GELU}
\left(
\mathbf{W}^{s}_{1}\mathbf{T}^{s}
\right)
\right),
\qquad
\widetilde{\mathbf{T}}^{s}
=
\mathbf{T}^{s}
\odot
(\textbf{1}+\eta_s\mathbf{g}^{s}), \nonumber
\end{equation*}
where $\mathbf{g}^{s}$ denotes the learned feature-wise modulation, and $\eta_s$ controls its strength. $\odot$ denotes element-wise multiplication, $\mathbf{1}$ is an all-ones vector with the same dimension as $\mathbf{T}^{s}$, and $\eta$ controls the strength of the gate-induced modulation. The gated representation is passed through the bottleneck transformation to provide a compact nonlinear adaptation space:
\begin{equation*}
\mathbf{h}^{s}
=
\operatorname{GELU}
\left(
\mathbf{W}^{s}_{\mathrm{down}}
\widetilde{\mathbf{T}}^{s}
\right),
\qquad
\mathbf{u}^{s}
=
\mathbf{W}^{s}_{\mathrm{up}}\mathbf{h}^{s}, \nonumber
\end{equation*}

Here, $\mathbf{W}^{s}_{\mathrm{down}}$ projects the gated representation into the bottleneck space, while $\mathbf{W}^{s}_{\mathrm{up}}$ maps the refined representation back to the target feature dimension. The bottleneck enables the visual representation to be adapted in a compact feature space before being mapped back to the target representation dimension.

The projected representation is then refined through a nonlinear residual transformation:
\begin{equation*}
\mathbf{z}^{s}
=
\operatorname{LN}
\left(
\mathbf{u}^{s}
+
\rho_s\mathcal{R}_s(\mathbf{u}^{s})
\right)
\tag{1}
\label{eq:ReGRAM_final_output}
\end{equation*}
where $\mathcal{R}_{s}(\cdot)$ denotes the refinement transformation, $\rho$ controls the contribution of the residual branch, and $\operatorname{LN}(\cdot)$ denotes layer normalization. The residual pathway allows the adapted representation to retain the information from the main projection while incorporating an additional nonlinear refinement. The resulting $\mathbf{z}^{s}$ is used as the adapted representation of visual stream $s$ for subsequent neural--visual alignment.

The adapted representations are finally fused using a learnable stream weight:
\begin{equation*}
\mathbf{z}_v
=
\sigma(\alpha)\mathbf{z}^{F}
+
\left(1-\sigma(\alpha)\right)\mathbf{z}^{M}  \nonumber
\end{equation*}
where $z^F$ and $z^M$ are the outputs of the proposed adapter module using Equation~\eqref{eq:ReGRAM_final_output}. The scalar $\alpha$ controls their relative contribution to the final visual representation.

\begin{algorithm*}[t]
\caption{Training and Protocol-specific Evaluation}
\label{alg:training}
\begin{algorithmic}[1]

\Statex \textbf{Input:} Paired neural--image dataset $\mathcal{D}$, pretrained
vision backbone $F_V$, total training epochs $T$, and $R$ snapshot intervals.
\Statex \textbf{Output:} Neural--image similarity matrix $\mathbf{S}$.

\State Initialize trainable parameters $\Theta$ and freeze the parameters of
$F_V$

\For{$e=1,\ldots,T$}
    \For{each paired training batch $(\mathbf{X},\mathbf{I})\subset\mathcal{D}$}
        \State Compute batch-level neural and visual representations
        $(\mathbf{Z}^{E},\mathbf{Z}^{V})$ using the proposed encoders
        \State Compute the bidirectional contrastive loss
        $\mathcal{L}_{\mathrm{CLIP}}(\mathbf{Z}^{E},\mathbf{Z}^{V})$
        \State Update $\Theta$ using AdamW
    \EndFor
    \State Update the learning rate using CAWR
\EndFor

\If{intra-subject protocol}
    \State Compute normalized test representations
    $(\bar{\mathbf{Z}}^{E},\bar{\mathbf{Z}}^{V})$ using the final model
    \State $\mathbf{S}\gets
    \bar{\mathbf{Z}}^{E}(\bar{\mathbf{Z}}^{V})^{\top}$
    \State Evaluate $\mathbf{S}$ using Top-1, Top-5, and mAP

\Else 
    \Statex \hspace{\algorithmicindent}\text{(Inter-subject protocol)}
    \State Divide the $T$ training epochs into $R$ snapshot intervals
    \For{$r=1,\ldots,R$}
        \State Select $\Theta_r$ with the highest validation Top-1 accuracy
        within snapshot interval $r$
        \State Compute normalized test representations
        $(\bar{\mathbf{Z}}^{E}_{r},\bar{\mathbf{Z}}^{V}_{r})$ using $\Theta_r$
        \State $\mathbf{S}_{r}\gets
        \bar{\mathbf{Z}}^{E}_{r}
        (\bar{\mathbf{Z}}^{V}_{r})^{\top}$
    \EndFor
    \State $\mathbf{S}\gets
    \frac{1}{R}\sum_{r=1}^{R}\mathbf{S}_{r}$
    \State Evaluate $\mathbf{S}$ using Top-1, Top-5, and mAP
\EndIf

\end{algorithmic}
\end{algorithm*}

\subsection{Cross-modal Representation Learning}

To align the neural and visual representations, we employ a symmetric bidirectional contrastive learning objective.  The similarity between the two modalities is computed using the scaled cosine similarity, where $\tau>0$ is the
temperature parameter controlling the concentration of the similarity
distribution.

The bidirectional contrastive loss is defined as
\begin{equation*}
\mathcal{L}_{\mathrm{CLIP}}
=
-\frac{1}{2B}
\sum_{i=1}^{B}
\Biggl(
\log
\frac{
\exp\left(
(\mathbf{z}_{e,i}^{\top}\mathbf{z}_{v,i})/\tau
\right)
}{
\displaystyle
\sum_{j=1}^{B}
\exp\left(
(\mathbf{z}_{e,i}^{\top}\mathbf{z}_{v,j})/\tau
\right)
}
+
\log
\frac{
\exp\left(
(\mathbf{z}_{v,i}^{\top}\mathbf{z}_{e,i})/\tau
\right)
}{
\displaystyle
\sum_{j=1}^{B}
\exp\left(
(\mathbf{z}_{v,i}^{\top}\mathbf{z}_{e,j})/\tau
\right)
}
\Biggr). \nonumber
\end{equation*}

The first term encourages each EEG representation to match its corresponding
visual representation, while the second performs the reverse image-to-EEG
alignment. The complete framework is optimized jointly using
$\mathcal{L}_{\mathrm{CLIP}}$, thereby learning a shared neural--visual
embedding space for zero-shot retrieval.

\section{Experiments}

We evaluate the proposed framework on the benchmark THINGS-EEG ~\cite{gifford2022large} and THINGS-MEG ~\cite{ds004212:2.0.0} under intra-subject and inter-subject settings. We first describe the experimental setup with respect to datasets and preprocessing procedures, followed by the experimental protocols and implementation details. We then present comparisons with existing approaches and ablation studies to examine the effect of the proposed design.

\subsection{Experimental Setup}

The experimental setup is designed to evaluate neural--visual retrieval across different neural modalities and subject-generalization settings. We consider both EEG and MEG data and evaluate the proposed framework under intra-subject and inter-subject protocols. The following subsections describe the datasets, preprocessing, evaluation protocols, and implementation details used in our experiments.

\begin{table}[H]
\centering
\caption{Top-1 and Top-5 retrieval accuracy (\%) for 200-way
zero-shot retrieval on the THINGS-EEG dataset.}
\label{tab:things_eeg_results}

\begingroup
\renewcommand{\arraystretch}{0.85}
\setlength{\tabcolsep}{3pt}

\resizebox{\textwidth}{!}{%
\begin{tabular}{@{}llccccccccccc@{}}
\toprule
\textbf{Method} & \textbf{Metric} &
\textbf{Sub1} & \textbf{Sub2} & \textbf{Sub3} &
\textbf{Sub4} & \textbf{Sub5} & \textbf{Sub6} &
\textbf{Sub7} & \textbf{Sub8} & \textbf{Sub9} &
\textbf{Sub10} & \textbf{Average} \\
\midrule

\multicolumn{13}{c}{\textbf{Intra-subject: Subject-Specific Evaluation}} \\
\midrule

\textbf{BraVL}~\cite{du2023decoding}
& Top-1 & 6.1 & 4.9 & 5.6 & 5.0 & 4.0 & 6.0 & 6.5 & 8.8 & 4.3 & 7.0 & 5.8 \\
& Top-5 & 17.9 & 14.9 & 17.4 & 15.1 & 13.4 & 18.2 & 20.4 & 23.7 & 14.0 & 19.7 & 17.5 \\
\midrule

\textbf{NICE}~\cite{song2024decoding}
& Top-1 & 12.3 & 10.4 & 13.1 & 16.4 & 8.0 & 14.1 & 15.2 & 20.0 & 13.3 & 14.9 & 13.8 \\
& Top-5 & 36.6 & 33.9 & 39.0 & 47.0 & 26.9 & 40.6 & 42.1 & 49.9 & 37.1 & 41.9 & 39.5 \\
\midrule

\textbf{UBP}~\cite{wu2025bridging}
& Top-1 & 41.2 & 51.2 & 51.2 & 51.1 & 42.2 & 57.5 & 49.0 & 58.6 & 45.1 & 61.5 & 50.9 \\
& Top-5 & 70.5 & 80.9 & 82.0 & 76.9 & 72.8 & 83.5 & 79.9 & 85.8 & 76.2 & 88.2 & 79.7 \\
\midrule

\textbf{ATS}~\cite{wu2026shrinking}
& Top-1 & 53.0 & 62.0 & 61.5 & 57.0 & 55.0 & 68.0 & 53.0 & 66.5 & 58.5 & 67.5 & 60.2 \\
& Top-5 & 79.0 & 87.5 & 89.0 & 86.5 & 84.0 & 90.5 & 84.0 & 91.0 & 86.0 & 89.0 & 86.7 \\
\midrule

\textbf{NeuroBridge}~\cite{zhang2026neurobridge}
& Top-1 & 50.0 & 63.2 & 61.6 & 61.4 & 54.8 & 69.7 & 62.7 & 71.2 & 64.0 & 73.6 & 63.2 \\
& Top-5 & 77.6 & 90.6 & 91.1 & 90.0 & 85.0 & 92.9 & 88.8 & 95.1 & 91.0 & 97.1 & 89.9 \\
\midrule

\textbf{BI-Cap}~\cite{shao2026braininspiredcaptureevidencedrivenneuromimetic}
& Top-1 & 65.0 & 77.0 & 74.5 & 69.5 & 69.0 & 76.5 & 65.5 & 77.5 & 72.5 & 76.5 & 72.4 \\
& Top-5 & 87.5 & 97.0 & 97.5 & 93.5 & 92.0 & 96.5 & 94.0 & 98.5 & 97.0 & 98.5 & 95.2 \\
\midrule

\textbf{Shallow Align}~\cite{du2026deep}
& Top-1 & 75.0 & 87.5 & 83.2 & 79.5 & 74.6 & 89.9 & 78.5 & 86.9 & 81.3 & 89.3 & 82.6 \\
& Top-5 & 94.3 & 98.9 & 98.2 & 96.1 & 96.4 & 99.3 & 97.3 & 99.4 & 97.8 & 99.1 & 97.7 \\
\midrule

\textbf{Ours}
& Top-1 & 92.3 & 96.0 & 93.0 & 88.7 & 90.4 & 93.9 & 89.9 & 95.6 & 91.4 & 93.8 & \textbf{92.5} \\
& Top-5 & 98.6 & 100.0 & 99.4 & 99.0 & 99.1 & 100.0 & 98.0 & 100.0 & 98.7 & 99.9 & \textbf{99.27} \\

\midrule
\multicolumn{13}{c}{\textbf{Inter-subject: Leave-One-Subject-Out Evaluation}} \\
\midrule

\textbf{BraVL}~\cite{du2023decoding}
& Top-1 & 2.3 & 1.5 & 1.4 & 1.7 & 1.5 & 1.8 & 2.1 & 2.2 & 1.6 & 2.3 & 1.8 \\
& Top-5 & 8.0 & 6.3 & 5.9 & 6.7 & 5.6 & 7.2 & 8.1 & 7.6 & 6.4 & 8.5 & 7.0 \\
\midrule

\textbf{NICE}~\cite{song2024decoding}
& Top-1 & 7.6 & 5.9 & 6.0 & 6.3 & 4.4 & 5.6 & 5.6 & 6.3 & 5.7 & 8.4 & 6.2 \\
& Top-5 & 22.8 & 20.5 & 22.3 & 20.7 & 18.3 & 22.2 & 19.7 & 22.0 & 17.6 & 28.3 & 21.4 \\
\midrule

\textbf{UBP}~\cite{wu2025bridging}
& Top-1 & 11.5 & 15.5 & 9.8 & 13.0 & 8.8 & 11.7 & 10.2 & 12.2 & 15.5 & 16.0 & 12.4 \\
& Top-5 & 29.7 & 40.0 & 27.0 & 32.3 & 33.8 & 31.0 & 23.8 & 32.2 & 40.5 & 43.5 & 33.4 \\
\midrule

\textbf{ATS}~\cite{wu2026shrinking}
& Top-1 & 17.0 & 21.5 & 7.5 & 13.5 & 7.0 & 14.5 & 10.0 & 13.0 & 9.5 & 26.0 & 14.0 \\
& Top-5 & 37.5 & 49.5 & 20.0 & 41.5 & 27.0 & 37.0 & 33.5 & 30.5 & 28.5 & 52.5 & 35.8 \\
\midrule

\textbf{NeuroBridge}~\cite{zhang2026neurobridge}
& Top-1 & 23.2 & 21.2 & 13.2 & 17.0 & 14.5 & 25.0 & 15.3 & 20.1 & 13.7 & 27.2 & 19.0 \\
& Top-5 & 52.4 & 49.3 & 36.5 & 45.3 & 37.7 & 55.0 & 45.1 & 44.9 & 36.5 & 56.3 & 45.9 \\
\midrule

\textbf{BI-Cap}~\cite{shao2026braininspiredcaptureevidencedrivenneuromimetic}
& Top-1 & 24.5 & 28.5 & 11.5 & 20.0 & 15.5 & 15.0 & 20.5 & 13.5 & 13.0 & 31.5 & 19.4 \\
& Top-5 & 54.5 & 62.0 & 29.5 & 48.5 & 41.5 & 44.5 & 41.5 & 37.0 & 36.0 & 65.0 & 46.0 \\
\midrule

\textbf{Shallow Align}~\cite{du2026deep}
& Top-1 & 23.5 & 30.6 & 10.0 & 19.5 & 18.1 & 22.7 & 18.6 & 17.3 & 23.0 & 34.4 & 21.8 \\
& Top-5 & 53.2 & 60.4 & 28.1 & 48.3 & 45.2 & 49.8 & 46.0 & 46.1 & 54.8 & 62.0 & 49.4 \\
\midrule

\textbf{Ours}
& Top-1 & 29.1 & 36.2 & 12.8 & 22.0 & 18.5 & 17.3 & 21.0 & 16.5 & 14.7 & 35.0 & \textbf{22.31} \\
& Top-5 & 61.8 & 65.0 & 32.2 & 54.8 & 45.6 & 45.1 & 51.4 & 39.0 & 35.6 & 68.3 & \textbf{49.88} \\

\bottomrule
\end{tabular}%
}
\endgroup
\end{table}

\subsubsection{Datasets}

\label{subsec1}
THINGS-EEG and THINGS-MEG are based on the THINGS ~\cite{hebart2023things} object concept database and follow a zero-shot protocol. Data is acquired under a rapid serial visual presentation paradigm, where each image is flashed for 100 ms, followed by a 100 ms blank screen, giving a 200 ms stimulus onset asynchrony. 

\medskip
\textbf{THINGS-EEG} ~\cite{gifford2022large}, a large-scale EEG-image dataset collected from ten participants using a 63-channel system. The standard zero-shot split yields 1,654 training concepts and 200 unseen test concepts, corresponding to 1654 concepts $\times$ 10 images $\times$ 4 training repetitions and 200 concepts $\times$ 1 image $\times$ 80 test repetitions to ensure reliable neural responses.

\medskip
\textbf{THINGS-MEG} ~\cite{ds004212:2.0.0} comprises high-density MEG recordings from four participants acquired with a 271-channel system. The dataset covers 1,854 training concepts (200 held-out concepts) under the zero-shot protocol. Training trials use 1654 concepts $\times$ 12 images $\times$ 1 repetition, and the test set follows 200 concepts $\times$ 1 image $\times$ 12 repetitions to obtain reliable estimates. Visual stimuli are presented for 500 ms, followed by a blank screen lasting 1000 ms$\pm$200 ms.

\subsubsection{Preprocessing}
\label{subsec1}
We follow a similar protocol to Wu et al. ~\cite{wu2025bridging}, with dataset-specific adaptations. For THINGS-EEG, the raw 63-channel recordings are band-pass filtered between 0.1 and 100 Hz. Epochs ran from stimulus onset to 1000 ms post-onset, and baseline correction is applied on the preceding 200 ms window. We downsampled the data from 1000 Hz to 250 Hz. Trial repetitions from the training and testing set are averaged to increase the signal-to-noise ratio, and multivariate noise normalization (MVNN) is applied per session to whiten the data. For the THINGS-MEG dataset, the 271 sensors are filtered within the same band, epoched identically, and downsampled to 200 Hz. We averaged the twelve test repetitions and applied channel-wise z-scoring across trials to increase the signal-to-noise ratio.

\subsubsection{Evaluation Metrics}

For a set of neural and image representations, the pairwise neural--image
similarity matrix is computed as
\begin{equation*}
\mathbf{S}
=
\bar{\mathbf{Z}}^{E}
(\bar{\mathbf{Z}}^{V})^{\top}, \nonumber
\end{equation*}
where $\bar{\mathbf{Z}}^{E}$ and $\bar{\mathbf{Z}}^{V}$ denote the
$\ell_2$-normalized neural and visual representations, respectively.
Each row of $\mathbf{S}$ corresponds to a neural query, and the columns
correspond to candidate images. The candidate images are ranked according
to their similarity scores.

We evaluate neural--image retrieval performance using Top-1 accuracy and Top-5 accuracy. Top-1 accuracy denotes the percentage of queries for which the corresponding target image is ranked first, while Top-5 accuracy indicates the percentage of queries in which the target image is retrieved within the top five ranked results.

\subsubsection{Experimental Protocol}

We evaluate the proposed framework under intra-subject and inter-subject conditions using the same training objective but different evaluation procedures. For both settings, the model is trained for $T$ epochs with cosine annealing with warm restarts (CAWR). For intra-subject evaluation, the model obtained at the final training epoch is directly used for test-time retrieval without checkpoint selection or snapshot ensembling. For inter-subject evaluation, the training process is divided into $R$ snapshot intervals. Within each interval, the checkpoint achieving the highest validation Top-1 accuracy is retained. Each retained checkpoint is evaluated independently on the test set to obtain normalized neural and visual embeddings corresponding to a similarity matrix. The resulting similarity matrices are averaged to obtain the final snapshot-level ensemble. Retrieval performance is reported using Top-1 accuracy and Top-5 accuracy.

\subsubsection{Implementation details}
The proposed framework is implemented in PyTorch and optimized using AdamW with an initial learning rate of $1\times10^{-4}$. CAWR periodically adjusts the
learning rate to encourage exploration of different model states during
optimization in an inter-subject experimentation setting. The model states are subsequently used for snapshot-based evaluation in the inter-subject setting. The neural branch employs DMSTAE, while the visual branch uses a frozen pretrained DINO-ViT-B/16 backbone with ReGRAM for hierarchical feature adaptation. Foveated and multi-scale DoG transformations are applied to the input images, and features from selected layers $1$, $3$, $5$, and $7$ are projected to a 768-dimensional space. All experiments are conducted on an NVIDIA RTX PRO 4500 Blackwell GPU machine.

\begin{table}[!htbp]
\centering
\caption{Top-1 and Top-5 retrieval accuracy (\%) on the
THINGS-MEG dataset under intra-subject and inter-subject
evaluation settings.}
\label{tab:things_meg_results}

\begingroup
\renewcommand{\arraystretch}{1.05}
\setlength{\tabcolsep}{3pt}

\resizebox{\linewidth}{!}{%
\begin{tabular}{@{}lcccccccccc@{}}
\toprule

\multirow{2}{*}{\textbf{Method}} &
\multicolumn{2}{c}{\textbf{Subject 1}} &
\multicolumn{2}{c}{\textbf{Subject 2}} &
\multicolumn{2}{c}{\textbf{Subject 3}} &
\multicolumn{2}{c}{\textbf{Subject 4}} &
\multicolumn{2}{c}{\textbf{Average}} \\

\cmidrule(lr){2-3}
\cmidrule(lr){4-5}
\cmidrule(lr){6-7}
\cmidrule(lr){8-9}
\cmidrule(lr){10-11}

& Top-1 & Top-5
& Top-1 & Top-5
& Top-1 & Top-5
& Top-1 & Top-5
& Top-1 & Top-5 \\
\midrule

\multicolumn{11}{c}{
\textbf{Intra-subject: Subject-Specific Evaluation}} \\
\midrule

\textbf{NICE}~\cite{song2024decoding}
& 9.6 & 27.8 & 18.5 & 47.8 & 14.2 & 41.6
& 9.0 & 26.6 & 12.8 & 36.0 \\

\textbf{UBP}~\cite{wu2025bridging}
& 15.0 & 38.0 & 46.0 & 80.5 & 27.3 & 59.0
& 18.5 & 43.5 & 26.7 & 55.2 \\

\textbf{ATS}~\cite{wu2026shrinking}
& 18.0 & 43.0 & 56.0 & 89.0 & 36.0 & 67.0
& 19.5 & 50.0 & 32.4 & 62.3 \\

\textbf{NeuroBridge}~\cite{zhang2026neurobridge}
& 16.5 & 41.6 & 53.7 & 85.3 & 40.4 & 73.2
& 18.1 & 43.1 & 32.2 & 60.8 \\

\textbf{BI-Cap}~\cite{shao2026braininspiredcaptureevidencedrivenneuromimetic}
& 19.0 & 45.5 & 70.0 & 94.0 & 48.5 & 82.0
& 24.0 & 54.5 & 40.4 & 69.0 \\

\textbf{Shallow Align}~\cite{du2026deep}
& 25.5 & 54.5 & 81.9 & 97.4 & 56.0 & 87.5
& 28.6 & 58.3 & 48.0 & 74.4 \\

\textbf{Ours}
& 29.4 & 56.1 & 82.8 & 97.1 & 53.0 & 82.5
& 35.8 & 64.5 & \textbf{50.25} & \textbf{75.05} \\

\midrule

\multicolumn{11}{c}{
\textbf{Inter-subject: Leave-One-Subject-Out Evaluation}} \\
\midrule

\textbf{UBP}~\cite{wu2025bridging}
& 2.0 & 5.7 & 1.5 & 17.2 & 2.7 & 10.5
& 2.5 & 8.0 & 2.2 & 10.4 \\

\textbf{ATS}~\cite{wu2026shrinking}
& 2.0 & 6.0 & 6.0 & 14.0 & 4.5 & 15.0
& 1.0 & 9.5 & 3.4 & 11.2 \\

\textbf{NeuroBridge}~\cite{zhang2026neurobridge}
& 4.3 & 13.1 & 3.6 & 15.6 & 3.0 & 11.2
& 2.5 & 11.3 & 3.4 & 12.8 \\

\textbf{BI-Cap}~\cite{shao2026braininspiredcaptureevidencedrivenneuromimetic}
& 2.5 & 6.0 & 6.0 & 22.0 & 5.0 & 16.5
& 2.0 & 9.5 & 3.9 & 13.5 \\

\textbf{Shallow Align}~\cite{du2026deep}
& 1.3 & 6.9 & 6.6 & 18.5 & 5.4 & 18.5
& 1.5 & 7.9 & 3.7 & 13.0 \\

\textbf{Ours}
& 3.0 & 9.0 & 4.2 & 20.6 & 7.5 & 22.7
& 4.1 & 13.5 & \textbf{4.7} & \textbf{16.45} \\

\bottomrule
\end{tabular}%
}

\endgroup
\end{table}

\subsection{Results and Analysis}

We evaluate the retrieval performance of the proposed framework across the two neural modalities and under intra-subject and inter-subject settings. The analysis focuses on Top-1 and Top-5 retrieval accuracy to assess the ability of the learned neural representations to identify the corresponding images in a zero-shot setting. Tables~\ref{tab:things_eeg_results} and~\ref{tab:things_meg_results} summarize the quantitative results for THINGS-EEG and THINGS-MEG, respectively.
On THINGS-EEG, intra-subject retrieval reaches $92.50\%$ Top-1 and $99.27\%$ Top-5 accuracy, while inter-subject retrieval achieves $22.31\%$ and $49.88\%$, respectively. On THINGS-MEG, the corresponding results are $50.25\%$ and $75.05\%$ for intra-subject retrieval and $4.70\%$ and $16.45\%$ for inter-subject retrieval.

\subsection{Ablation Study}

\subsubsection{Robustness to Representation and Adaptation Choices}

\begin{figure*}
    \centering
    \begin{subfigure}[t]{0.48\textwidth}
        \centering
        \includegraphics[width=\textwidth]{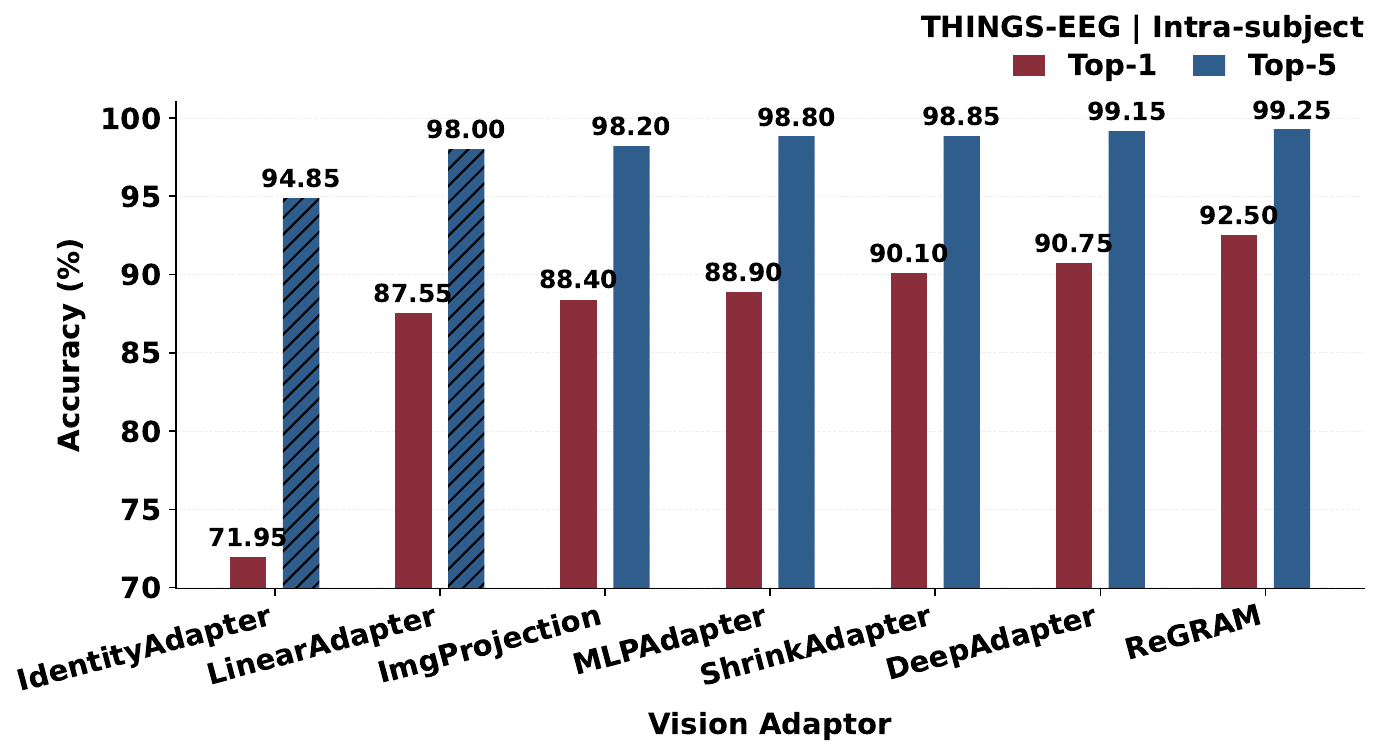}
        \caption{}
        \label{fig:vision_adapter_ablation}
    \end{subfigure}
    \hfill
    \begin{subfigure}[t]{0.48\textwidth}
        \centering
        \includegraphics[width=\textwidth]{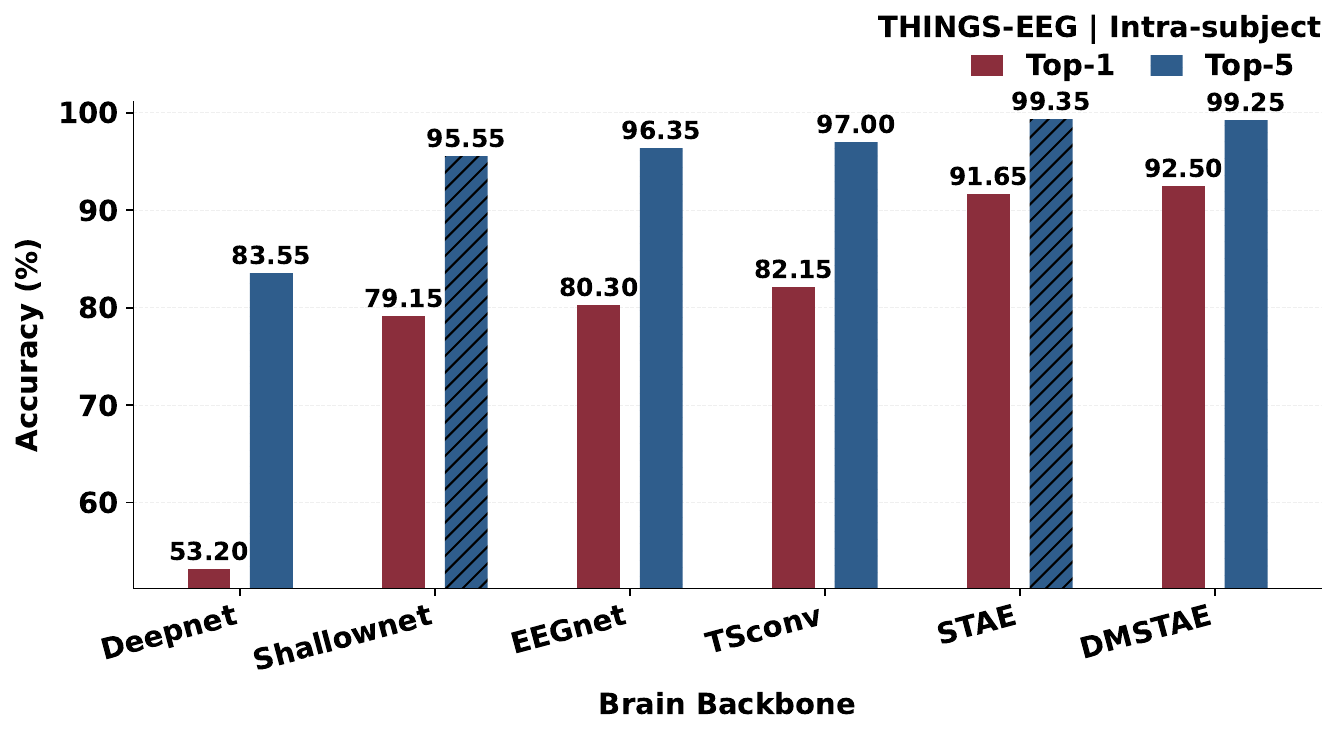}
        \caption{}
        \label{fig:brain_encoder_ablation}
    \end{subfigure}

    \caption{Ablation study of the model components (a) Comparison of different vision adapter architectures (b) Comparison of different brain encoder architectures.}
    \label{fig:ablation_comparison}
\end{figure*}

To assess whether the effectiveness of the proposed framework depends on specific architectural choices, we evaluate different visual adapters and neural encoders under the intra-subject setting. For the visual branch, the ablation examines whether adapting the pretrained visual representation is beneficial and whether the proposed ReGRAM provides an advantage over alternative adaptation strategies. As shown in Figure~\ref{fig:ablation_comparison}(a), ReGRAM achieves the best performance, reaching 92.5\% Top-1 and 99.25\% Top-5 accuracy, indicating the proposed gated residual refinement provides an effective way to adapt hierarchical visual features for neural--visual alignment. For the neural branch, we investigate how different brain encoder architectures affect the quality of the learned neural representation. As shown in Figure~\ref{fig:ablation_comparison}(b), temporal-attention-based encoders outperform conventional alternatives, with DMSTAE achieving the best overall retrieval performance.
This indicates that explicitly modeling temporal dependencies in the neural signals is important for obtaining representations that are more suitable for cross-modal alignment. 

\subsubsection{Benefit of Complementary Biologically Guided Visual Representations}

A single visual representation may not capture all the information relevant to neural responses. To examine whether different biologically guided representations provide complementary information, we compare foveated and multi-scale DoG transformations individually and in combination with direct and uniform representations.

\begin{figure*}
    \centering
    \includegraphics[width=0.72\textwidth]{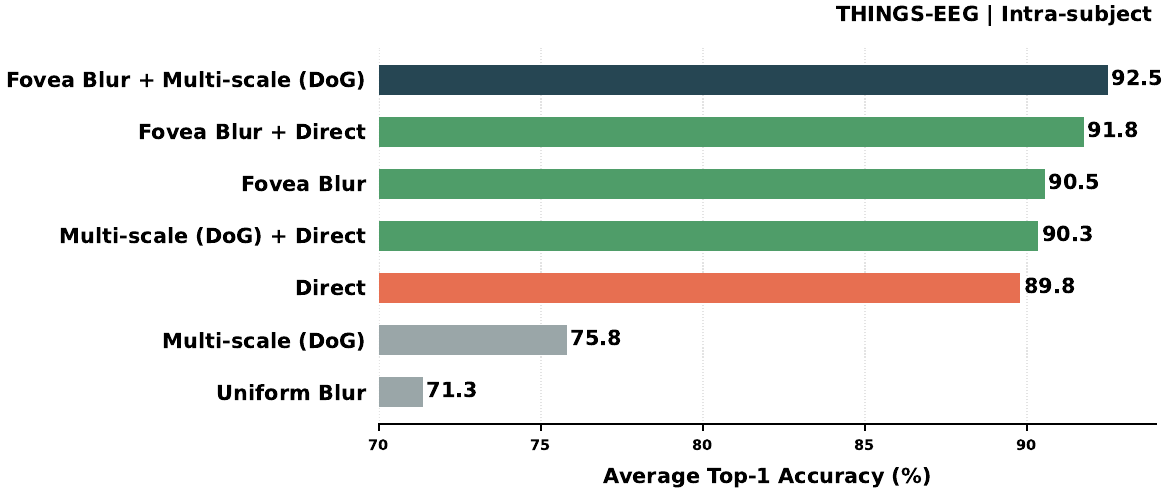}
    \caption{Intra-subject retrieval performance under different image transformation strategies.}
    \label{fig:ablation on image transformation}
\end{figure*}

The foveated and multi-scale DoG streams individually outperform the Direct and Uniform representations, indicating that explicitly modeling structured spatial resolution and luminance-driven contrast provides visual information that is better aligned with neural responses.  More importantly, combining the two streams achieves the highest retrieval performance, exceeding either individual representation. As shown in Figure~\ref{fig:ablation on image transformation}, combining the foveated and multi-scale DoG representations achieves the highest retrieval performance.  This improvement suggests that the two transformations capture complementary aspects of visual information rather than redundant cues: the foveated stream emphasizes spatially selective visual structure, while the multi-scale DoG stream emphasizes luminance contrast and coarse spatial information. The combination provides a richer visual representation for matching the diverse information encoded in neural responses, supporting the use of multiple biologically guided visual streams rather than relying on a single visual transformation.

\subsubsection{Hierarchical Visual Representations Strengthen Alignment}

The visual backbone encodes information at different levels of the
visual hierarchy, and we investigate whether neural--visual alignment benefits
from combining representations across multiple depths rather than relying on a single layer or a fixed stage of the backbone. 
\begin{figure*}
    \centering
    \includegraphics[width=0.72\textwidth]{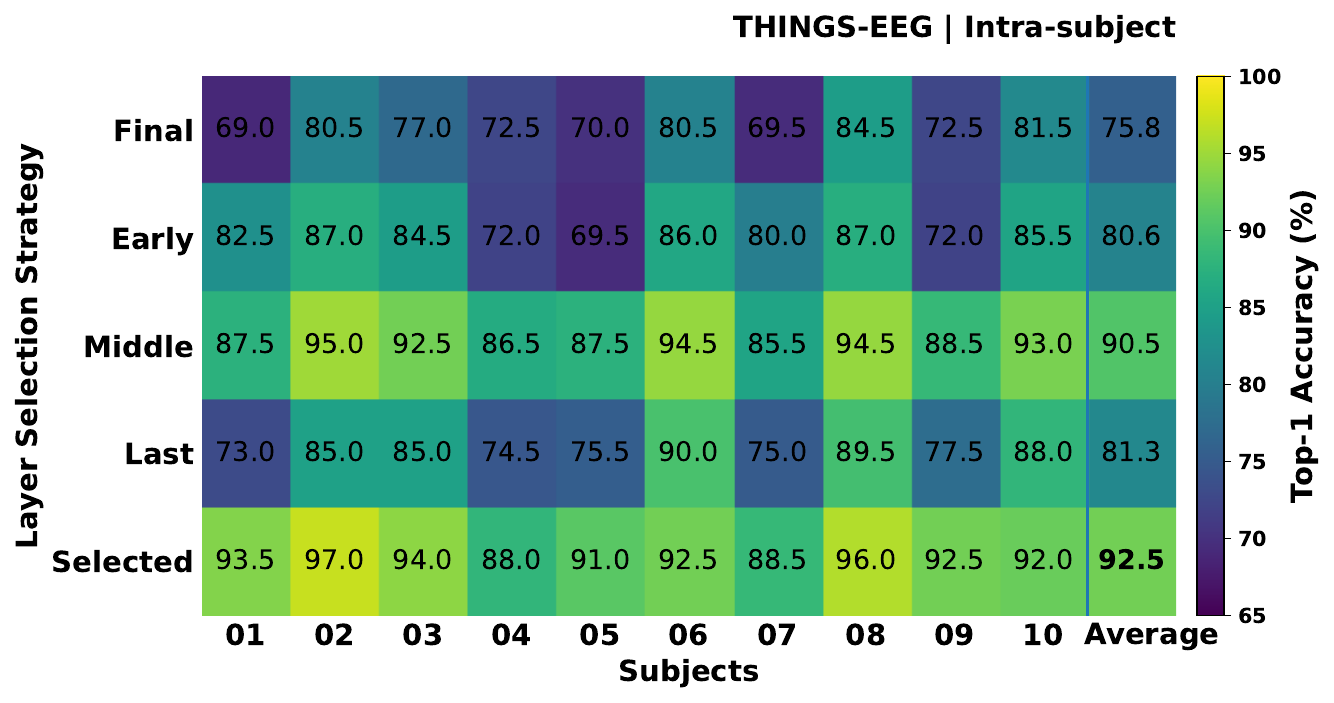}
    \caption{layer selection heatmap }
    \label{fig:Layer selection}
\end{figure*}
The layer-selection results in Figure~\ref{fig:Layer selection} show that the selected-layer strategy achieves the highest average Top-1 accuracy of 92.5\%, consistently outperforming the four early, four middle, four last, and final-layer alternatives. This improvement suggests that different visual depths provide complementary information for neural--visual alignment, with intermediate representations retaining lower-level visual structure while deeper representations capture increasingly abstract information. The results support the use of adaptive layer
selection to preserve complementary information across the visual hierarchy. 

\subsubsection{Progressive Contribution of Framework Components}

The proposed framework introduces several components that operate at different stages of the neural--visual alignment pipeline. While individual ablations can reveal the effect of a particular component, they do not show how the components collectively contribute to the final representation. We therefore perform a progressive ablation in which the framework is constructed incrementally from a neural baseline by successively introducing hierarchical visual representations, biologically guided visual streams, ReGRAM, and the residual connection. Thus, we hypothesize that progressively incorporating these components will lead to corresponding improvement in the framework. To quantify these contributions, each component is cumulatively added to the preceding configuration while keeping the remaining experimental settings unchanged. Table~\ref{tab:progressive_ablation} reports the resulting Top-1 and Top-5 retrieval accuracies, together with the absolute gain over the preceding configuration.
\begin{table*}
\centering
\caption{Progressive contribution of the framework components on the THINGS-EEG dataset. Each component is cumulatively added to the preceding configuration. Gain denotes the absolute improvement over the previous configuration in percentage points.}
\label{tab:progressive_ablation}
\begin{tabular}{lcccc}
\hline
Configuration & Top-1 & Gain & Top-5 & Gain \\
\hline
DMSTAE & 62.65 & -- & 85.90 & -- \\
+ Intermediate layers & 72.85 & +10.20 & 95.25 & +9.35 \\
+ Fovea Blur + Multi-scale (DoG) & 75.35 & +2.50 & 95.05 & -0.20 \\
+ ReGRAM & 89.85 & +14.50 & 99.00 & +3.95 \\
+ Residual connection & 92.50 & +2.65 & 99.25 & +0.25 \\
\hline
\end{tabular}
\end{table*}

Progressive results show that successive components contribute substantially to retrieval performance, although their effects differ between stages, with the complete framework achieving 92.50\% Top-1 and 99.25\% Top-5 accuracy. Intermediate-layer representations provide a substantial initial gain of 10.20\% in Top-1 accuracy, while adding the foveated and multi-scale DoG streams further improves performance. ReGRAM yields the largest additional gain of 14.50\%, highlighting the importance of effective visual feature adaptation, while the residual connection provides an additional 2.65\% improvement.

\subsubsection{Robustness of Evaluation Strategy}

The evaluation protocol can influence the stability of retrieval performance, particularly when the learned neural representations are sensitive to the training state. This sensitivity may be more pronounced in inter-subject retrieval, where the model must generalize across subject-specific variations in neural responses. We therefore examine whether the choice of model state used for evaluation affects retrieval performance under different training conditions.

\begin{figure*}
    \centering

    \centering
    \begin{subfigure}[t]{0.47\textwidth}
        \centering
        \includegraphics[width=\linewidth]{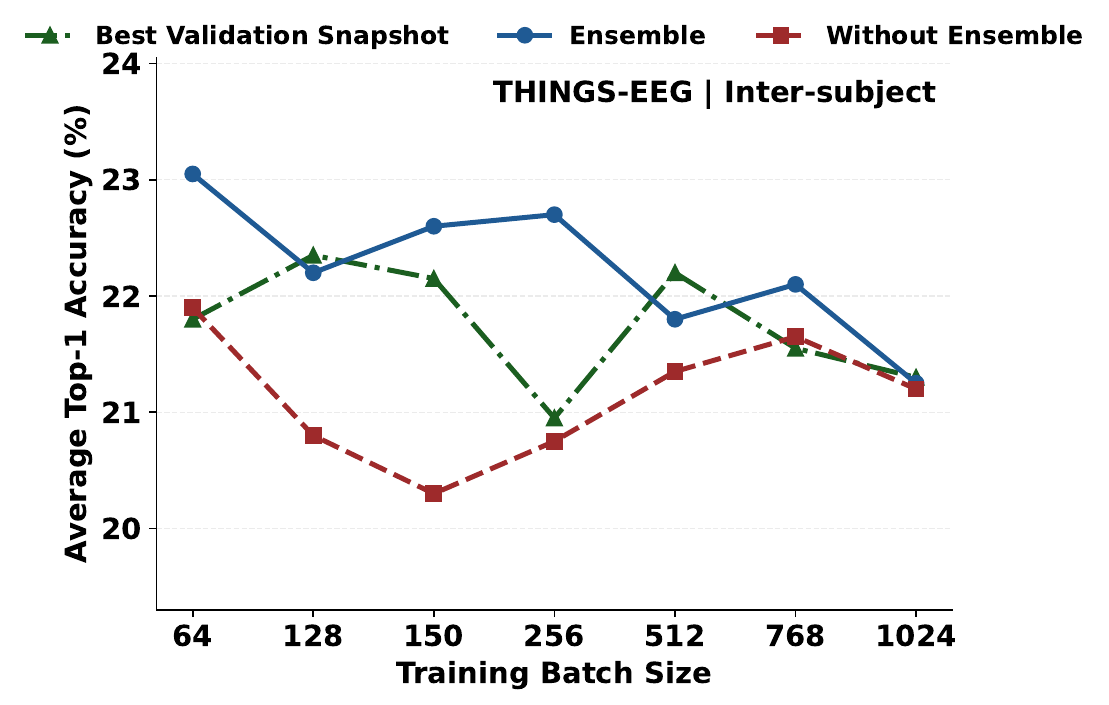}
        \caption{}
        \label{fig:batch_eeg_inter}
    \end{subfigure}
    \hfill
    \begin{subfigure}[t]{0.47\textwidth}
        \centering
        \includegraphics[width=\linewidth]{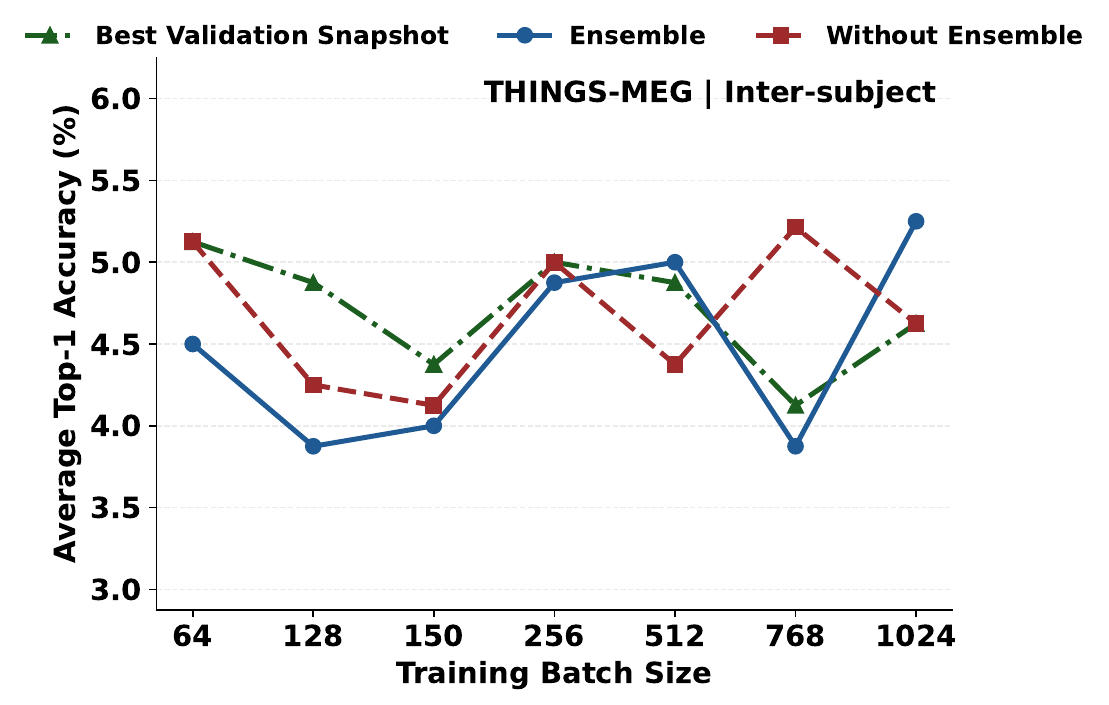}
        \caption{}
        \label{fig:batch_meg_inter}
    \end{subfigure}
    \caption{Comparison of best-validation-snapshot, snapshot-ensemble, and  without ensemble evaluation across training batch sizes inter-subject retrieval on THINGS-EEG and THINGS-MEG.}
    \label{fig:batch_size_inter}
\label{fig:evaluation strategy}
\end{figure*}

To examine the robustness of the evaluation strategy across training conditions and subject settings, we compare the final model, snapshot ensemble, and best-validation-snapshot strategies across different batch sizes. The results in Figure~\ref{fig:evaluation strategy} shows that the ensembling remains competitive across
training batch sizes.
In the more challenging inter-subject setting, snapshot-based strategies generally provide more stable and stronger retrieval performance, with the snapshot ensemble performing particularly well across the evaluated batch sizes. This suggests that aggregating representations from multiple training snapshots reduces the sensitivity to a single model state and provides a more robust representation for
cross-subject neural--visual alignment.

\subsubsection{Qualitative Retrieval Analysis}

\begin{figure*}
    \centering
    \begin{subfigure}[t]{0.48\textwidth}
        \centering
        \includegraphics[width=\linewidth]{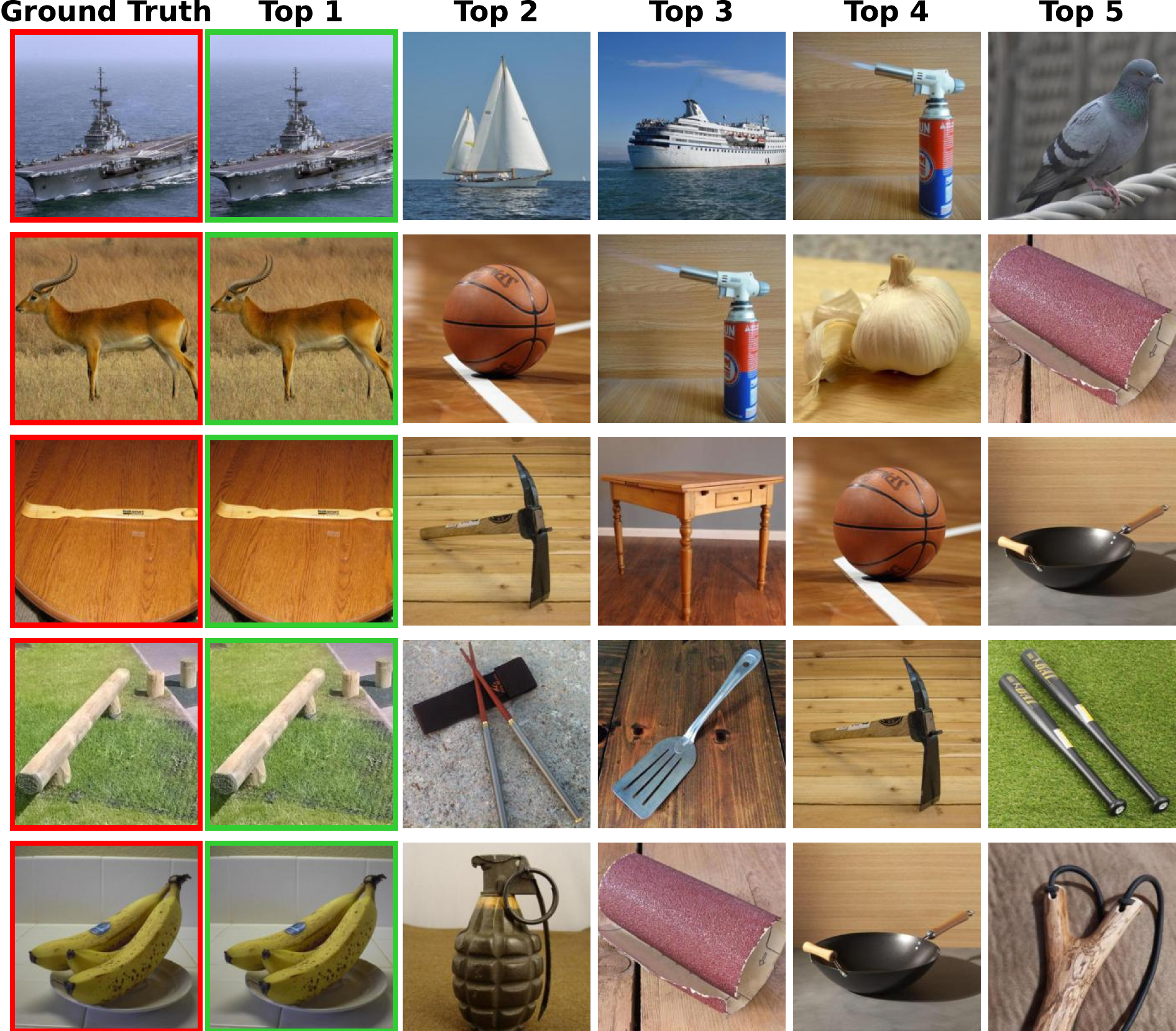}
        \caption{Successful retrieval cases}
        \label{fig:retrieval_best}
    \end{subfigure}
    \hfill
    \begin{subfigure}[t]{0.48\textwidth}
        \centering
        \includegraphics[width=\linewidth]{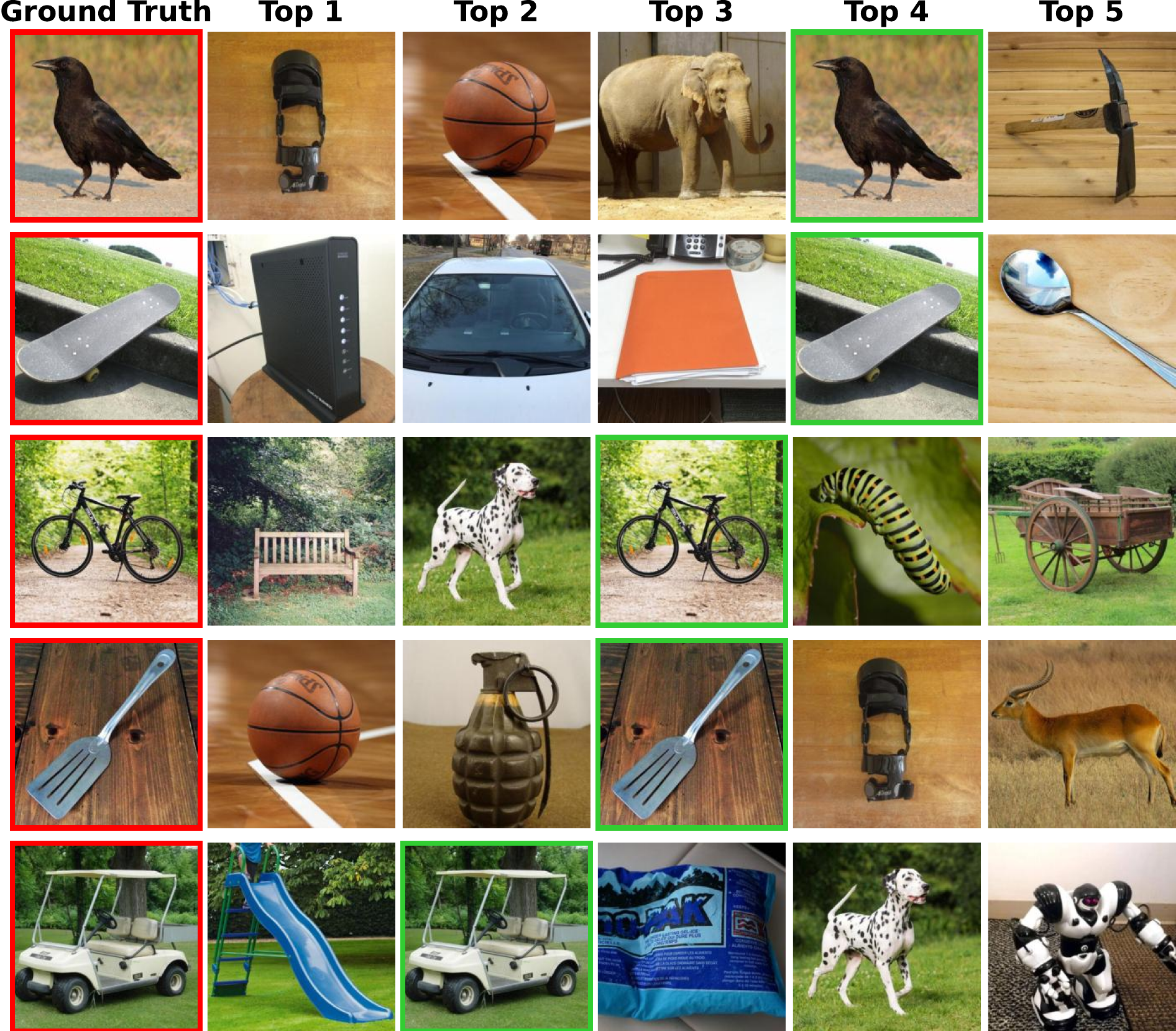}
        \caption{Challenging retrieval cases}
        \label{fig:retrieval_worst}
    \end{subfigure}
    \caption{Representative retrieval results on the THINGS-EEG dataset, showing successful and challenging retrieval cases.}
    \label{fig:qualitative_retrieval}
\end{figure*}

Quantitative retrieval metrics summarize overall performance; however, they hide how the learned representation behaves for individual retrieval queries. To examine the quality of the retrieved images, we visualize representative successful and challenging cases from the THINGS-EEG dataset in Figure~\ref{fig:qualitative_retrieval}.

 The successful cases show that the learned neural representations retrieve visually related target images with high similarity, while the challenging cases illustrate instances where retrieval becomes less discriminative. These examples provide qualitative evidence that the learned cross-modal representation captures meaningful visual information from neural responses, while also highlighting the difficult retrieval cases. Interestingly, incorrect retrievals are often visually related to the target, suggesting that the learned representation preserves meaningful perceptual similarities even when the exact image is not retrieved.

\section{Conclusion}
Neural–visual decoding is challenging because neural responses and visual representations have different structures, temporal dynamics, and levels of abstraction. This becomes particularly evident when a model is required to retrieve previously unseen visual concepts or generalize across subjects. We present neural--visual representation learning framework for zero-shot visual retrieval from neural signals. The framework integrates direction-aware temporal modeling with biologically guided visual streams, hierarchical visual representations, and adaptive feature refinement to establish a more effective neural--visual alignment. The results provide a broader indication that effective neural--visual alignment depends not only on the final embedding space, but also on how the information reaching that space is represented and refined. However, the difference between intra-subject and inter-subject performance suggest that subject-specific neural variability remains an important factor in the learned representations. Future work can therefore investigate mechanisms for reducing this variability and learning more transferable neural representations, to cross-subject generalization in zero-shot neural--visual decoding.



\bibliographystyle{cas-model2-names}
\bibliography{cas-refs}
\end{document}